\pdfoutput=1

\documentclass[11pt]{article}

\usepackage[preprint]{acl}

\usepackage{times}
\usepackage{latexsym}
\usepackage{multirow}
\usepackage[T1]{fontenc}

\usepackage[utf8]{inputenc}

\usepackage{microtype}

\usepackage{inconsolata}

\usepackage{graphicx}
\usepackage{subcaption}

\usepackage{comment}
\usepackage{tcolorbox}
\usepackage{amsmath,amssymb}
\usepackage{booktabs}
\usepackage{xcolor,caption,subcaption,graphicx}
\usepackage{tabularx,multirow,float}

\title{\textsc{Hoaxpedia}: A Unified Wikipedia Hoax Articles Dataset}

\author{
\textbf{Hsuvas Borkakoty\textsuperscript{1}},
\textbf{Luis Espinosa-Anke\textsuperscript{1,2}}
\\
\textsuperscript{1}Cardiff NLP, School of Computer Science and Informatics, Cardiff University, UK 
\\
\textsuperscript{2}AMPLYFI, UK
\\
\small{
   \texttt{\{borkakotyh,espinosaankel\}@cardiff.ac.uk}
 }
}

\begin{document}
\maketitle
\begin{abstract}
Hoaxes are a recognised form of disinformation created deliberately, with potential serious implications in the credibility of reference knowledge resources such as Wikipedia. What makes detecting Wikipedia hoaxes hard is that they often are written according to the official style guidelines. In this work, we first provide a systematic analysis of similarities and discrepancies between legitimate and hoax Wikipedia articles, and introduce \textsc{Hoaxpedia}, a collection of 311 hoax articles (from existing literature and official Wikipedia lists), together with semantically similar legitimate articles, which together form a binary text classification dataset aimed at fostering research in automated hoax detection. In this paper, We report results after analyzing several language models, hoax-to-legit ratios, and the amount of text classifiers are exposed to (full article vs the article's definition alone). Our results suggest that detecting deceitful content in Wikipedia based on content alone is hard but feasible, and complement our analysis with a study on the differences in distributions in edit histories, and find that looking at this feature yields better classification results than context. \footnote{The Dataset is available at: \url{https://huggingface.co/datasets/hsuvaskakoty/hoaxpedia} and associated codes are available at: \url{https://github.com/hsuvas/hoaxpedia\_dataset.git} } 
\end{abstract}

\section{Introduction}
\label{sec:intro}
Wikipedia is, as \citet{hovy2013collaboratively} defines it, the ``largest and most popular collaborative and multilingual resource of world and linguistic knowledge'', and it is acknowledged that its accuracy is on par with or superior than, e.g., the Encyclopedia Britannica \cite{giles2005special}. However, as with any other platform, Wikipedia is also the target of online vandalism, and \textit{hoaxes}, a more obscure, less obvious form of vandalism\footnote{\url{https://en.wikipedia.org/wiki/Wikipedia:Do_not_create_hoaxes}.}, constitute a significant threat to Wikipedia's overall integrity \citep{kumar2016disinformation,wong2021wikireliability,wangmckeown2010got}, among others, because of its ``publish first, ask questions later'' policy \cite{asthana2018with}. Although Wikipedia employs community based New Page Patrol systems to check the credibility of a newly created article, the process is always in backlog\footnote{\url{https://en.wikipedia.org/wiki/Wikipedia:New_pages_patrol}.}, making it overwhelming \cite{schinder2014accept}.
\begin{figure}[t]
            \centering
            \includegraphics[width=1\linewidth]{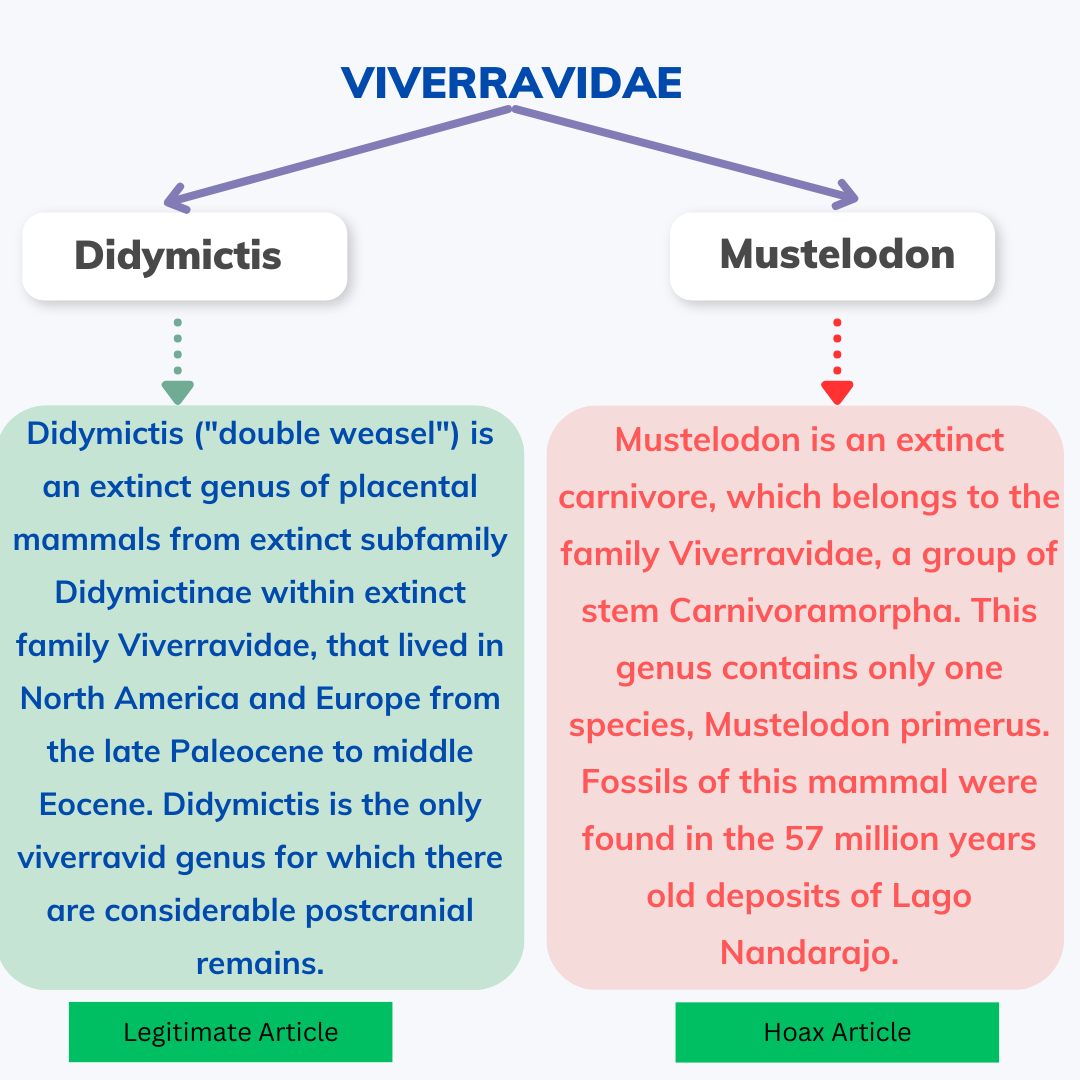}
            \caption{Example of Hoax and Legitimate(Real) article}
            \label{fig:example_hoax}
        \end{figure}


Hoax articles (as shown in Figure \ref{fig:example_hoax}), are created to deliberately spread false information \cite{kumar2016disinformation}, harm the credibility of Wikipedia as a knowledge resource and generate concerns among its users \cite{hu2007measuring}. Since manual inspection of quality is a process typically in backlog \cite{dang2016quality}, the automatic detection of such articles is a desirable feature. However, most works in the literature have centered their efforts in metadata associated with hoax articles, e.g., user activity, appearance features or revision history \cite{zeng2006trust,elebiary2023role, kumar2016disinformation,wong2021wikireliability,hu2007measuring, susuri2017detection}. For example, \citet{adler2011wikipedia} introduced a vandalism detection system using metadata, content and author reputation features, whereas \citet{kumar2016disinformation} provide a comprehensive study of hoax articles and their timeline from discovery to deletion. In their work, the authors define the characteristics of a successful hoax, with a data-driven approach based on studying a dataset of 64 articles (both hoax and legitimate), on top of which they train statistical classifiers. Furthermore, other works have compared network traffic and features of hoax articles to those of other articles published the same day \citep{elebiary2023role}, and conclude that hoax articles attract more attention after creation than \textit{cohort} (or legitimate) articles. Finally, \citet{wong2021wikireliability} study various Wikipedia vandalism types and introduce the Wiki-Reliability dataset, which comprises articles based on 41 author-compiled templates. This dataset contains 1,300 articles marked as hoax, which are legitimate articles with false information, a.k.a hoax facts \citep{kumar2016disinformation}.

In this paper, we propose to incorporate content to the mix of features/signals that are fed to classifiers for the hoax detection task, and aim to answer the research question. 
We first construct a dataset (\textsc{HoaxPedia}) containing 311 hoax articles and around 30,000 \textit{plausible negative examples}, i.e., legitimate Wikipedia articles that are semantically similar to hoax articles, to create a set of negatives that \textit{cover similar topics} to hoax articles (e.g., a newly discovered species). We also explore whether a Wikipedia definition (the first sentence of the article) can provide hints towards its veracity. Our results (reported at different ratios of hoax vs. legitimate articles) suggest that, while style and shallow features are certainly not good the best predictors, language models (LMs) are capable of exploiting other more intricate features (e.g., an article's revision history) and open a promising research direction focused on content-based hoax flagging.  Our contributions in this work can be summarised as follows.
\begin{itemize}
    \item We systematically contrast a set of proven Wikipedia hoax articles with legitimate articles. 
    \item We propose \textsc{HoaxPedia}, a novel Wikipedia Hoax article dataset with 311 hoax articles and 30,000 semantically similar legitimate articles collected from Wikipedia.
    \item We conduct binary classification experiments on HoaxPedia, using a range of language models (including LLMs), features and hoax-to-legitimate ratio.
\end{itemize}

\section{Related work}
\label{sec:lit_review}
In what follows, we give a brief overview of disinformation detection, the datasets available for the community and the role of Wikipedia in disinformation detection, as our work falls in the intersection between disinformation detection and Wikipedia research.

\paragraph{\noindent\textbf{Disinformation detection and datasets:}}Disinformation and misinformation are two types of false information, they differ in that misinformation is inaccurate information created or propagated unknowingly, whereas disinformation is inaccurate information deliberately created to mislead the intended consumer \cite{hernon1995disinformation,fallis2014functional,kumar2016disinformation,ireton2018journalism}. Nonetheless, both are harmful towards information quality and reliability, thus posing risks toward different aspects of society \cite{su2020motivations}. \citet{alam2021survey} surveys disinformation detection from a multi-modal perspective, showing that it can come in different modalities like text, images, audio and video, with text being the most common. Datasets used for disinformation detection can be divided based on the length of input or claim: short sentences (such as tweets or Reddit posts) vs articles (common type being news articles), where most of the datasets follow claim-evidence based format \cite{su2020motivations}. The short sentences or claim based datasets are mostly sourced from social media, such as X (formerly Twitter) \cite{castillo2011information, derczynski2017semeval, zubiaga2019detection,lopez2021focused}, Reddit \cite{gorrell2018rumoureval,qu2022disinfomeme}, or fact checking websites like Politifact\footnote{\url{https://www.politifact.com/}} \cite{wang2017liar}, Snopes\footnote{\url{https://www.snopes.com/}} \cite{vo2020facts}, or a combination of different fact checking websites \cite{augenstein2019multifc}. These datasets contain claims which need to be verified as well as the response, in the form of True/False labeling backed by evidences, which are also provided in these datasets. Article level datasets, on the other hand, are varied, and focus on state-backed propaganda \cite{heppell2023analysing}, German multi-label disinformation (the GerDISDETECT dataset) \cite{schutz2024gerdisdetect}, or narratives at conflict dataset containing news articles \cite{sinelnikhovy2024narratives}, which mostly focuses on news article or propaganda based disinformation spreading. The datasets mentioned above are specialized towards topic/trend based or news based disinformation, lacking a generalised aspect of disinformation detection with free text. 

\paragraph{\noindent\textbf{Wikipedia in disinformation detection:}} Wikipedia, as described by \citet{zachary2020it}, serves as a source of information validation as backed by its large set of articles contributed by community. This is seen in action for fact verification task datasets such as FEVER \cite{thorne2018fact}, TabFactA \cite{chen2019tabfact}, FNC-1 (Fake News Challenge-1) dataset \cite{pomerleau2017fake} etc. In these datasets, evidences for claims are collected from Wikipedia articles (eg. FEVER, FNC-1) and tables (eg. TabFactA). However, being a product of community effort, Wikipedia is also prone to vandalism and inaccurate contents \cite{zachary2020it}, and the community outlines different policies to combat these issues\footnote{\url{https://en.wikipedia.org/wiki/Wikipedia:Vandalism}}. We also find efforts to automatize the process of detecting vandalism contents from Natural Language Processing perspective. Preliminary approaches  include using feature based approaches extracted from metadata and editor behaviour to detect vandalism \cite{wu2010elusive,javanmardi2011vandalism,heindorf2016vandalism}. Implementation of early warning system based on metadata and editor behavior is found in work of \citet{kumar2015vews}, where they propose a dataset of page metadata with list of blocked user and unblocked users, and proposes a feature based matrix for classification using auto-encoders. \citet{yuan2017wikipedia} proposes an edit history based approach, where they use behaviour of users over time as feature to create the embedding space for multi-source LSTM network \cite{hochreiter1997long}. However, these approaches do not consider article text as a marker to detect the vandalism. Wikipedia marks hoax articles as form of vandalism \cite{kumar2016disinformation,thorne2018fever}. A comprehensive discussion about different approaches for hoax detection are mentioned in Section \ref{sec:intro}. However, it is worth noting that existing research on this topic is not sufficient to make meaningful progress towards effective disinformation detection using Wikipedia, most of which can be attributed to unavailability of a dedicated dataset containing hoax articles as well as their metadata. We find a research gap towards developing datasets and tools for hoax detection where article contents are used, with foundation stone laid by works of \citet{kumar2016disinformation} and \citet{wong2021wikireliability}. Through our work, we aim to address this by introducing \textsc{Hoaxpedia}, where we unify the existing datasets we found based on article text, as well as collect articles from Wikipedia's list of hoaxes.

\section{\textsc{HoaxPedia} Construction}
\label{sec:dataset}
\textsc{HoaxPedia} is constructed by unifying five different resources that contain known hoaxes, e.g., from \citet{kumar2016disinformation,elebiary2023role}, as well as the official Wikipedia hoaxes list\footnote{\url{https://en.wikipedia.org/wiki/Wikipedia:List_of_hoaxes_on_Wikipedia}} and the Internet Archive. We used Internet Archive to manually retrieve Wikipedia pages that are now deleted from Wikipedia, but were at one point in the past identified as hoaxes. We manually verify each of the article we collect from Wikipedia and Internet Archive as hoax using their accompanied deletion discussion and reasons for citing them as a hoax.  
In terms of negative examples, while we could have randomly sampled Wikipedia pages, this could have introduced a number of biases in the dataset, e.g., hoax articles contain historical events, personalities or artifacts, and thus we are interested in capturing a similar breadth of topics, entities and sectors in the negative examples so that a classifier cannot rely on these spurious features. We select these negative examples by ensuring they correspond to authentic content. This is achieved by verifying they do not carry the {{Db-hoax}} flag, which Wikipedia's New Page Patrol policy uses to mark potential hoaxes. Within this set, we extract negative examples as follows. Let $H$ be the set of hoax articles, and $W$ the set of candidate \textit{legitimate} Wikipedia pages, with $T_H = \lbrace t_{H^{1}}, \dots, t_{H^{p}} \rbrace$ and $T_W = \lbrace t_{W^{1}}, \dots, t_{W^{q}} \rbrace$ their corresponding vector representations, and $p$ and $q$ the number of hoax and candidate Wikipedia articles, respectively. Then, for each SBERT (all-MiniLM-L6-v2) \citep{reimers2019sentence} title embedding $t_{H^i} \in T_H$, we retrieve its top $k$ nearest neighbors (\textsc{NN}) from $T_W$ via cosine similarity \textsc{cos}. We experiment with different values for $k$, specifically $k \in \lbrace 2, 10, 100 \rbrace$:

\[
\mbox{\textsc{NN}}\left(t_{H^i}\right) = \lbrace t_{W^j} : j \in J_k(t_{H^i}) \rbrace 
\]

where $J_k(t_{H^i})$ contains the top $k$ cosine similarities in $T_W$ for a given $t_{H^i}$, and 

\[
\mbox{\textsc{cos}}\left(t_{H^i}, t_{W^j}\right) = \frac{t_{H^i} \cdot t_{W^j}}{\vert\vert t_{H^i} \vert\vert \vert\vert t_{W^j} \vert\vert}
\]

The result of this process is a set of positive (hoax) articles and a set of negative examples we argue will be similar in content and topic, effectively removing any topic bias from the dataset.

\section{Hoax vs. legitimate, a Surface-Level Comparison}

To maintain longevity and avoid detection, hoax articles follow Wikipedia guidelines and article structure. This raises the following question: \emph{``how (dis)similar are hoaxes with respect to a hypothetical legitimate counterpart?''}. 
Upon inspection, we found comments in the deletion discussions such as \emph{``I wouldn't have questioned it had I come across it organically''} (for the hoax article \textit{The Heat is On} \footnote{\url{https://en.wikipedia.org/wiki/Wikipedia:Articles_for_deletion/The_Heat_Is_On_(TV_series)}}), or \emph{``The story may have a "credible feel'' to it, but it lacks any sources''}, a comment on article \textit{Chu Chi Zui}\footnote{\url{https://en.wikipedia.org/wiki/Wikipedia:Articles_for_deletion/Chu_Chi_Zui}}. Comments like these highlight that hoaxes are generally well written (following Wikipedia's guidelines), and so we proceed to quantify their stylistic differences in a comparative analysis that looks at: (1) article text length comparison; (2) sentence and word length comparison; and (3) a readability analysis. 

\begin{figure}[ht]
            \centering
            \includegraphics[width=1\linewidth]{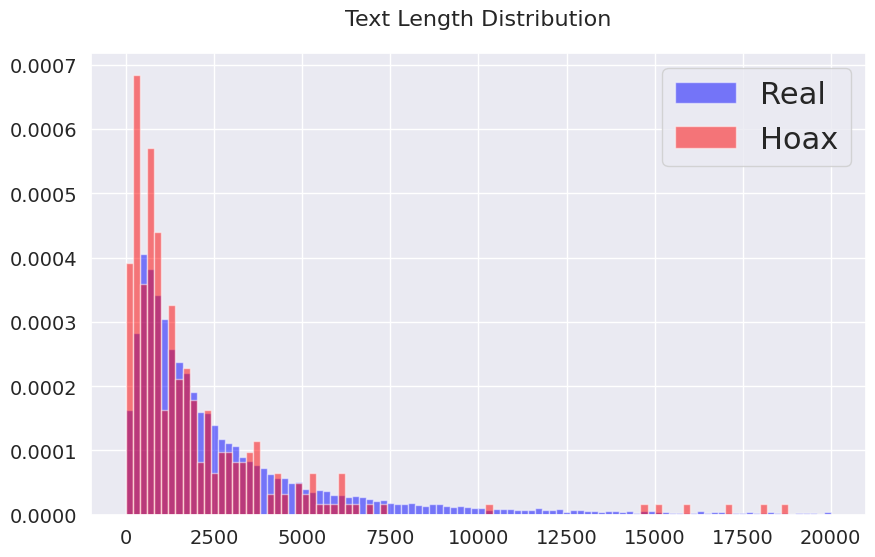}
            \caption{Text length distribution for hoax and legitimate articles.}
            \label{fig:text_len}
        \end{figure}

\noindent\paragraph{\textbf{Article Text length distribution:}} Following the works of \citet{kumar2016disinformation}, we conduct a text length distribution analysis with hoax and legitimate articles, and verify they show a similar pattern (as shown in Figure \ref{fig:text_len}), with similar medians for hoax and legitimate articles, specifically 1,057 and 1,777 words, respectively. 


\begin{figure*}
    \centering
    \begin{subfigure}{.33\linewidth}
        \centering
        \includegraphics[width=\linewidth]{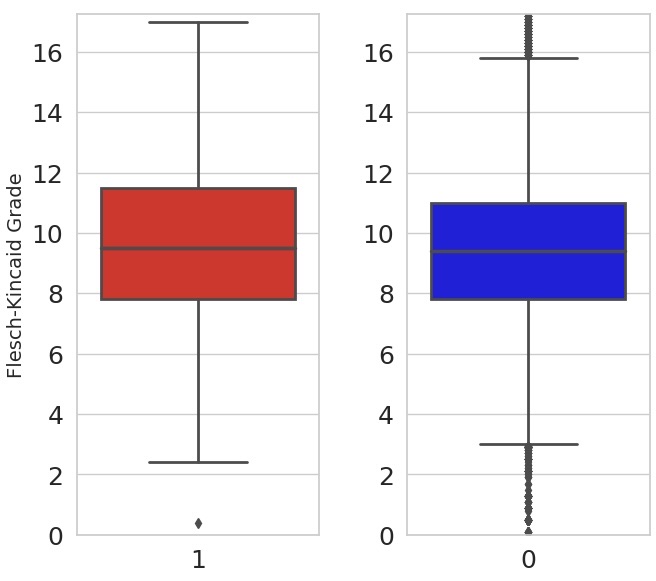}
        \subcaption{Flesch-Kincaid Grade}
        \label{fig:fk_grade}
    \end{subfigure}\hfill
    \begin{subfigure}{.33\linewidth}
        \centering
        \includegraphics[width=\linewidth]{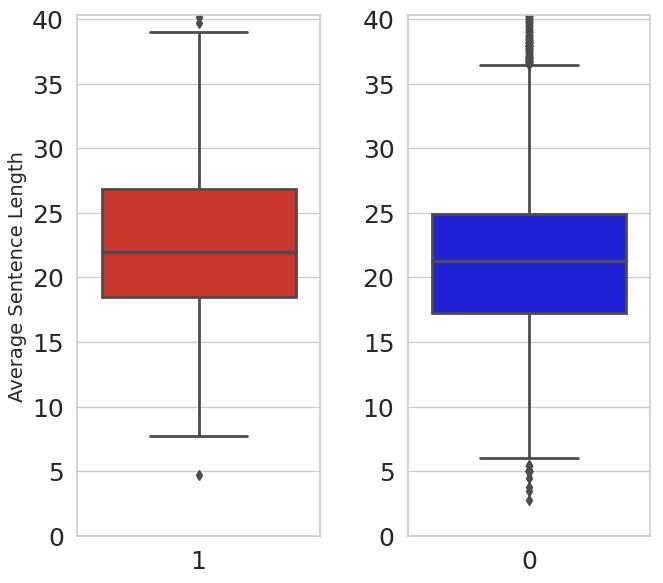}
        \subcaption{Average Sentence Length}
        \label{fig:average_sent_len}
    \end{subfigure}\hfill
    \begin{subfigure}{.33\linewidth}
        \centering
        \includegraphics[width=\linewidth]{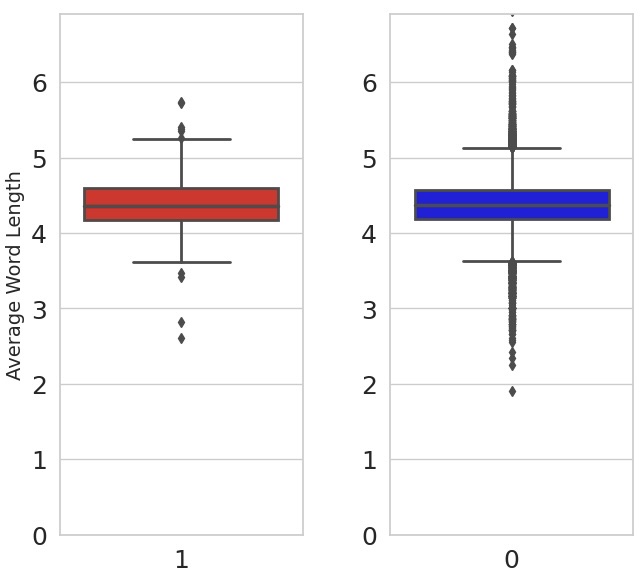}
        \subcaption{Average Word Length}
        \label{fig:average_word_len}
    \end{subfigure}
    \caption{Results of different stylistic analyses on Hoax (red) and legitimate (blue) articles.}
    \label{fig:all_style_analysis}
\end{figure*}

\noindent\paragraph{\textbf{Average sentence and word length:}} Calculating average sentence and word length for hoaxes and legitimate articles separately can be a valuable proxy for identifying any obvious stylistic or linguistic (e.g., syntactic complexity) patterns. We visualize these in a series of box plots in Figure \ref{fig:all_style_analysis}. They clearly show a similar style, with  sentence and word length medians at 21.23 and 22.0, and 4.36 and 4.35 for legitimate and hoax articles respectively. 

\noindent\paragraph{\textbf{Readability Analysis:}} Readability analysis gives a quantifiable measure of the complexities of texts. It can clarify on easement of understanding the text, revealing patterns that can be used to either disguise disinformation like hoaxes or convey clear, factual content. For readability analysis, we use Flesch-Kincaid (FK) Grading system \citep{flesch2007flesch}, a metric that indicates the comprehension difficulty when reading a passage in the context of contemporary academic English. This metric gives us an aggregate of the complexity of documents, their sentences and words, measured by the average number of words per sentence and the average number of syllables per word. After obtaining an average for both hoax and legitimate articles, we visualize these averages again in Figure \ref{fig:all_style_analysis}, we find a median of 9.4 for legitimate articles and 9.5 for hoax articles, again highlighting the similarities between these articles.

\begin{figure*}
    \centering
    \begin{subfigure}{0.49\textwidth} 
        \centering
        \includegraphics[width=\textwidth]{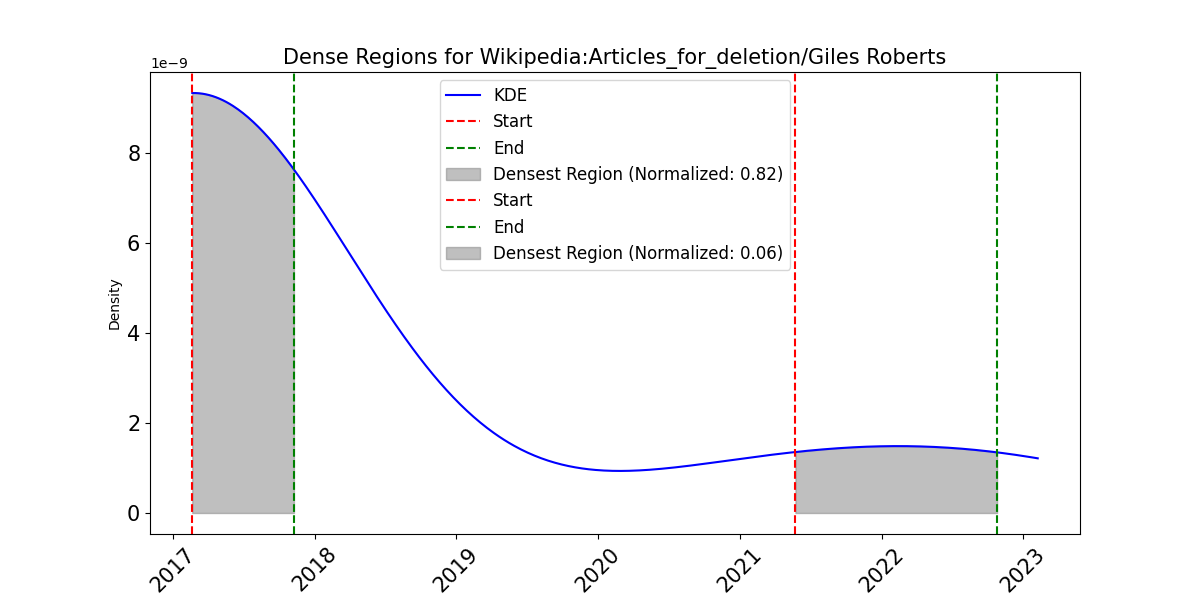}
        \subcaption{Plot for an example Hoax article }
        \label{fig:hoax_tl}
    \end{subfigure}
    \hfill  
    \begin{subfigure}{0.49\textwidth}  
        \centering
        \includegraphics[width=\textwidth]{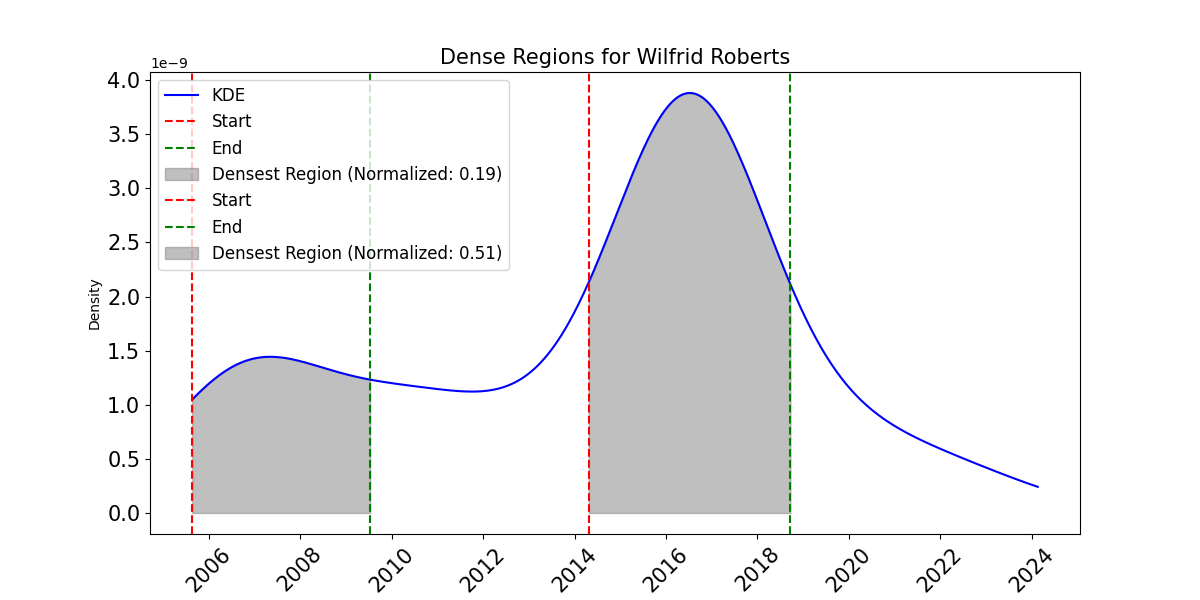}
        \subcaption{Plot for an example legitimate article }
        \label{fig:real_tl}
    \end{subfigure}
    
    \caption{Timeline based dense region plots for hoax and legitimate articles (with start and end lines for each region and normalized densities marked in grey)}
    \label{fig:timeline_plots}
\end{figure*}

\begin{figure}[ht]
        \centering
        \includegraphics[width=1\linewidth]{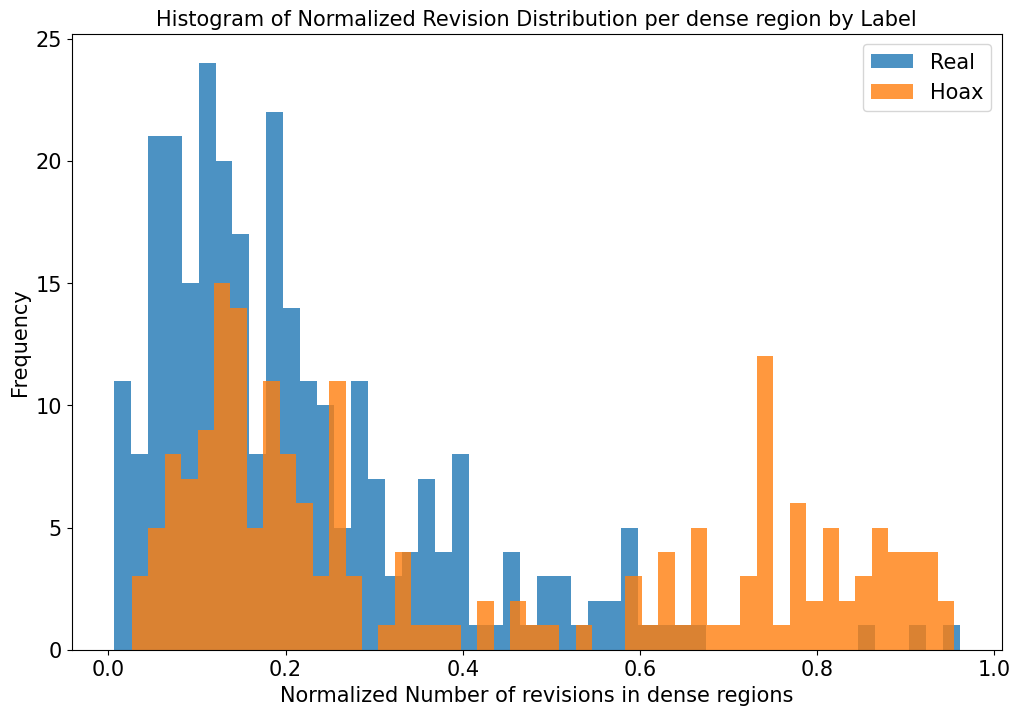}
        \caption{Histogram of normalized distribution for number of revisions in dense regions for hoax and legitimate (real) article}
        \label{fig:histogram_timeline}
    \end{figure}

\noindent\paragraph{\textbf{Timeline Analysis:}} Analysing the revision timelines of hoaxes and legitimate Wikipedia articles can give valuable insights into revision density and pattern that might differentiate the both sets of articles. To that end, we collect the timelines of hoax and legitimate articles (in all three hoax to legitimate ratio mentioned above) and add a timeline dataset to \textsc{Hoaxpedia}. Since some of the hoax articles were deleted from Wikipedia, so we are only able to obtain 164 hoax articles out of 311 in our dataset. With these hoax articles and their semantically similar legitimate counterparts, we construct the timeline dataset of \textsc{Hoaxpedia}.  We check the timeline patterns through the use of Bayesian Online Changepoint Detection (BOCPD) \cite{adams2007bayesian} with Kernel Density Estimation (KDE) \cite{wkeglarczyk2018kernel} to obtain the dense regions for revision histories of hoax and legitimate articles. Figure \ref{fig:timeline_plots} show a comparative timelines with marked normalized dense region for a hoax and a legitimate article as example. Timelines of hoax article show that they are quite dormant in terms of their number of revisions and their dense regions are mostly around the beginning and end of the article's timeline, which can be attributed to the new page patrol and deletion discussion of hoaxes respectively. To test this hypothesis further, we divide the revision timelines of hoax and legitimate articles into quartiles (divide the timeline into four equal parts) and count the number of dense regions over total dense regions. The result for each quartile is given in Table \ref{tab:quartile_distribution}, which shows that the proportion of dense region happening at the beginning and at the end are much higher for hoax articles than legitimate ones. We also show the normalised histogram distribution of hoax and legitimate (real) articles in Figure \ref{fig:histogram_timeline}. 

\begin{table}[htbp!]
\scriptsize 
\centering
\resizebox{0.6\columnwidth}{!}{%
\begin{tabular}{lrr}
\hline
\textbf{Quartile} & \multicolumn{1}{l}{\textbf{Hoax}} & \multicolumn{1}{l}{\textbf{Real}} \\ \hline
Q1                & 0.69                                        & 0.75                                        \\
Q2                & 0.02                                        & 0.17                                        \\
Q3                & 0.04                                        & 0.22                                        \\
Q4                & 0.75                                        & 0.42                                        \\ \hline
\end{tabular}%
}
\caption{Average distribution of dense regions per quartile (timeline divided into four parts) for hoax and legitimate (real) articles}
\label{tab:quartile_distribution}
\end{table}

Since we find difference in timeline patterns for hoax and legitimate articles, we also test the possibility of classifying hoax article using timeline only. We train a Support Vector Machine (SVM) \cite{vapnik2013nature} for this purpose, with TF-IDF \cite{sparck1972statistical} vectorizer fit on revision history binned by months (all revision timestamps converted to bins of MM-YYYY format) to build the timeline vocabulary. The results for Label 1 (Hoax) in F1-score is in Table \ref{tab:timeline_classifier}, which shows that the classifier is able to perform well with discerning hoax article from legitimate ones, with an effect of data imbalance clearly showing in model's performance. This further proves that revision history between hoaxes and legitimate articles are different in its pattern, and can be an important feature in detection of hoaxes. However, we also argue that timeline alone may not be able solve the detection of hoaxes, as it is a statistical feature prone to outliers.

\begin{table}[htbp!]
\centering
\resizebox{0.7\columnwidth}{!}{%
\begin{tabular}{lrrr}
\hline
\multicolumn{1}{c}{\textbf{Data Split}} & \multicolumn{1}{c}{\textbf{Precision}} & \multicolumn{1}{c}{\textbf{Recall}} & \multicolumn{1}{c}{\textbf{F1}} \\ \hline
1H2R                                    & 0.86                                   & 0.91                                & 0.88                            \\
1H10R                                   & 0.89                                   & 0.78                                & 0.83                            \\
1H100R                                  & 0.97                                   & 0.69                                & 0.80                            \\ \hline
\end{tabular}%
}
\caption{Results of SVM Timeline classifier for Label 1 (Hoax) for all data splits}
\label{tab:timeline_classifier}
\end{table}

The above analysis suggests that hoax Wikipedia articles are indeed well disguised, and so in the next section we propose a suite of experiments for hoax detection based on language models, setting an initial set of baselines for this novel dataset.

\section{Experiments}
\label{expt}
We cast the problem of identifying hoax vs. legitimate articles as a binary classification problem, in whcih we evaluate a suite of LMs, specifically: BERT family of models (BERT base and large \cite{devlin2019bert}, RoBERTa-base and large \cite{liu2019roberta}, Albert-base and large \cite{lan2019albert}), as well as T5 (Base and Large) \cite{raffel2020exploring} and Longformer (Base) \cite{beltagy2020longformer}. We use the same training configuration for the BERT-family of models, T5 models and Longformer, and set the generation objective as \emph{Binary classification} for the T5 models. In terms of data size, we consider the three different scenarios outlined in Section \ref{sec:dataset} (2x, 10x and 100x negative examples). This approach naturally increases the challenge for the classifiers. The details about the data used in different settings are given in Appendix \ref{app:dataset_details}.


In addition to the three different settings for positive vs negative ratios, we also explore \textit{how much text is actually needed to catch a hoax}, or in other words, \textit{are definition sentences in hoax articles giving something away}? This is explored by running our experiments on the full Wikipedia articles, on one hand, and on the definition (first sentence alone), on the other. This latter setting is interesting from a lexicographic perspective because it helps us understand if the Wikipedia definitions show any pattern that a model could exploit. Moreover, from the practical point of view of building a classifier that could dynamically \emph{``patrol''} Wikipedia and flag content automatically, a definition-only model would be more interpretable (with reduced ambiguity and focusing on core meaning/properties of the entity) and could have less parameters (handling smaller vocabularies, and compressed knowledge), which would have practical retraining/deployment implications in cost and turnaround. 

\section{Results}
\label{sec:result}
Our experiments are aimed to explore the impact of data imbalance and content length. In both cases, we compare several classifiers and analyze whether model size (in number of parameters) is correlated with performance. In terms of evaluation metrics, all results we report are F1 on the positive class (hoax). In definition-only setting, we find that models evaluated on datasets that are relatively balanced (2 real articles for every hoax) show a stable performance, but they degrade drastically as the imbalance increases. In terms of robustness, RoBERTa is the most consistent, with an F1 of around 0.6 for all three settings, whereas Albert models perform poorly (exhibiting, however, some interesting behaviours, which we will discuss further). For the full text setting, we find that Longformer models performs well, with an F1 of 0.8. Surprisingly, the largest model we evaluated (T5-large) is not the best performing model, although this could point to under-fitting, the dataset potentially being too small for a model this size to train properly. Another interesting behaviour of T5-large is that in the 1 hoax vs 2 real setting, it shows exactly the same performance, whether seeing a definition or the full text. On the other side of the spectrum, we find that Albert models are the ones showing the highest improvement when going from definition to full text. This is interesting, as it shows a small model may miss nuances in definitions but can still compete with, or even outperform, larger models.

\begin{table*}[]
\centering
\resizebox{\textwidth}{!}{%
\begin{tabular}{l>{\raggedleft\arraybackslash}p{2cm}>{\raggedleft\arraybackslash}p{2cm}>{\raggedleft\arraybackslash}p{2cm}>{\raggedleft\arraybackslash}p{2cm}>{\raggedleft\arraybackslash}p{2cm}>{\raggedleft\arraybackslash}p{2cm}>{\raggedleft\arraybackslash}p{2cm}}
\hline
\multicolumn{1}{c}{\textbf{}} & \textbf{}                               & \multicolumn{3}{c}{\textbf{Definition}}                                                                                                  & \multicolumn{3}{c}{\textbf{Fulltext}}                                                                                                   \\ \hline
\textbf{Model}                & \textbf{Model Size}                     & \textbf{1H2R}                     & \textbf{1H10R}                     & \textbf{1H100R}                    & \textbf{1H2R}                     & \textbf{1H10R}                     & \textbf{1H100R}                     \\ \hline
Albert-base-v2                & 12M                                     & 0.23                                       & 0.17                                        & 0.06                                        & 0.67                                       & 0.47                                        & 0.11                                         \\
Albert-large-v2               & 18M                                     & 0.28                                       & 0.30                                        & 0.15                                        & 0.72                                       & 0.63                                        & 0.30                                         \\
BERT-base                     & 110M                                    & 0.42                                       & 0.30                                        & 0.14                                        & 0.55                                       & 0.57                                        & 0.32                                         \\
RoBERTa Base                  & 123M                                    & 0.57                                       & 0.59                                        & 0.53                                        & 0.82                                       & 0.75                                        & 0.63                                         \\
Longformer-base               & 149M                                    & 0.43                                       & 0.35                                        & 0.54                                        & 0.80                                       & 0.78                                        & 0.67                                         \\
T5-Base                       & 220M                                    & 0.48                                       & 0.25                                        & 0.14                                        & 0.51                                       & 0.27                                        & 0.23                                         \\
BERT-large                    & 340M                                    & 0.43                                       & 0.36                                        & 0.17                                        & 0.61                                       & 0.64                                        & 0.33                                         \\
RoBERTa-large                 & 354M                                    & 0.58                                       & 0.63                                        & 0.62                                        & 0.84                                       & 0.81                                        & 0.79                                         \\
T5-large                      & 770M                                    & 0.54                                       & 0.32                                        & 0.13                                        & 0.54                                       & 0.43                                        & 0.37                                         \\ \hline
\end{tabular}%
}
\caption{F1 on the positive class - \textit{hoax} at different degrees of data imbalance for Definition and Fulltext Settings (H: Hoax, R: Real)}
\label{tab:all_setting_comparison}
\end{table*}

A perhaps not too surprising observation is that all models improve after being exposed to more text, as seen in Table \ref{tab:all_setting_comparison}, increasing their F1 about 20\% on average, and sometimes even up to 30. This confirms that definitions alone are not a sufficiently strong signal for detecting hoax articles, although there are notable exceptions. 
Moreover, in terms of absolute performance, RoBERTa models perform decently, although significantly below their full text settings. It is interesting to note that Longformer base yields much better results in the hardest setting (1 Hoax vs 100 Real) when exposed only to definitions. This is indeed a surprising and counter intuitive result that deserves future investigation.

\subsection{Ablation: Effect of Definitions on Classifying Hoaxes}
We run a data ablation test on the full-text split of the dataset to find the impact of definition sentences. To this end, we remove the first sentence of each article, and run the fulltext classification experiment, focusing on RoBERTa-Large, the most consistent model. The results of this ablation experiment are shown in Table \ref{tab:ablation}, suggesting that F1 goes down about 2\% for the positive class when the definition sentence is missing. This shows that definitions show critical information about entities and events in Wikipedia, but often are not the place where hoax features would emerge, and therefore removing them from the full text doesn't change much the story.

\begin{table}[]
\resizebox{\columnwidth}{!}{%
\begin{tabular}{lrrrr}
\hline
\multicolumn{1}{c}{\textbf{Model}} & \multicolumn{1}{c}{\textbf{Setting}} & \multicolumn{1}{c}{\textbf{Precision}} & \multicolumn{1}{c}{\textbf{Recall}} & \multicolumn{1}{c}{\textbf{F1}} \\ \hline
RoBERTaL                           & 1H2R                                 & 0.83                                   & 0.80                                & 0.82                            \\
RoBERTaL                           & 1H10R                                & 0.82                                   & 0.71                                & 0.76                            \\
RoBERTaL                           & 1H100R                               & 0.67                                   & 0.51                                & 0.58                            \\ \hline
\end{tabular}%
}
\caption{Performance of RoBERTa-Large on Binary Classifications Without Definition Sentences in Articles (with Fulltext hoax to real ratio in Settings column)}
\label{tab:ablation}
\end{table}

\section{Large Language Models and HoaxPedia}
\label{sec:result_llm}
We explore the capabilities of open source Large Language Models (LLM) in detecting hoax articles through our proposed dataset. We select Llama2-7B and 13B \cite{touvron2023llama}, Llama3-8B \cite{dubey2024llama}, and Mistral-7B models \cite{jiang2023mistral} for the experiments, and the prompts used are given in the Appendix \ref{app:prompt_llm}. We take three experimental settings: perplexity, instruction-tuning, and supervised fine-tuning. The settings and their results are described below.

\begin{figure*}
    \centering
    \begin{subfigure}{0.49\textwidth} 
        \centering
        \includegraphics[width=\textwidth]{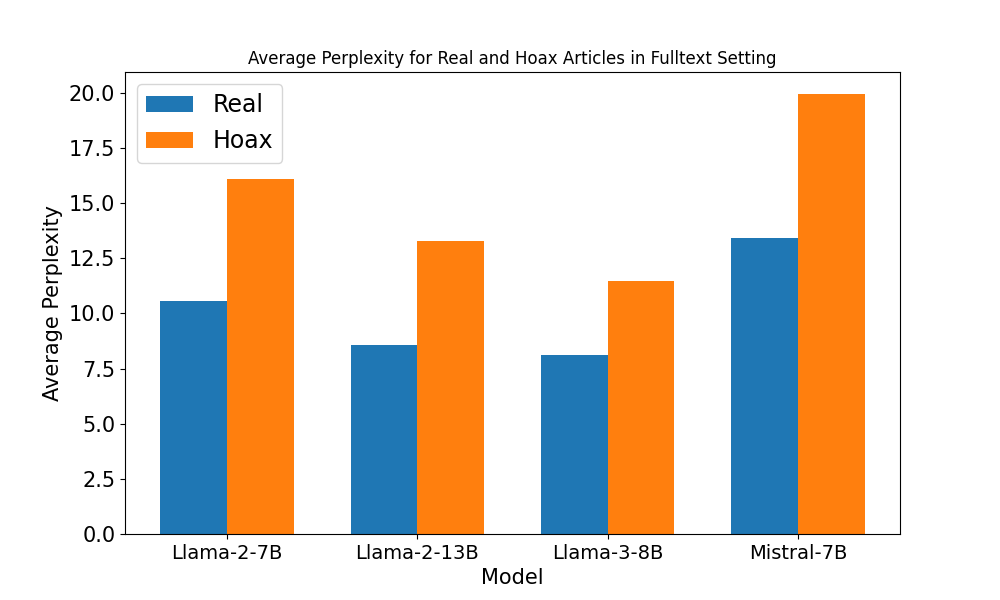}
        \subcaption{Average perplexity scores for fulltext}
        \label{fig:perplexity_ft}
    \end{subfigure}
    \hfill  
    \begin{subfigure}{0.49\textwidth}  
        \centering
        \includegraphics[width=\textwidth]{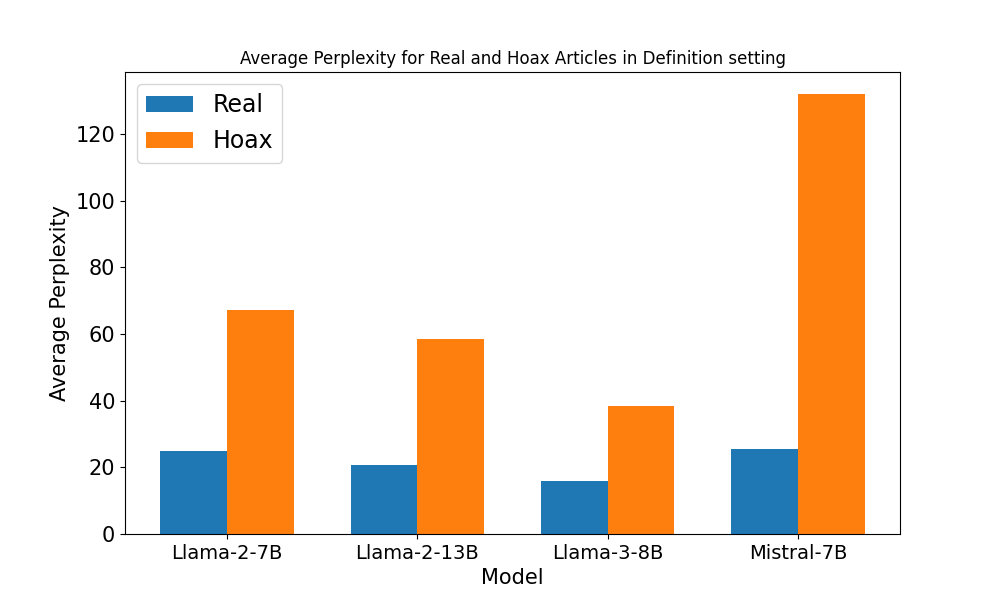}
        \subcaption{Average perplexity scores for definition}
        \label{fig:perplexity_def}
    \end{subfigure}
    
    \caption{Average Model based perplexity scores for Fulltext and Definition Settings}
    \label{fig:perplexity}
\end{figure*}

\noindent\textbf{Perplexity Experiments: }We consider perplexity as an indicator for LLMs to predict distribution of of Hoax and legitimate articles, with the hypothesis that LLMs will have difficulty for predicting contents of hoax articles, resulting in higher perplexity. We test the LLMs in both definition and fulltext settings. The average perplexity results for both settings are shown in Figure \ref{fig:perplexity}, revealing that there is a significant difference between perplexity of hoax and legitimate articles in both settings. This suggests that LLMs struggle to predict distribution of Hoax articles.  

\noindent\textbf{Prompting: }For prompting, we consider zero-shot and few-shot prompt settings. The prompts are given in Appendix \ref{app:prompt_llm}, and the input setting are for both definition and fulltext. We report the results for Macro-F1 scores. The results are reported in Table \ref{tab:f1_scores_instruct}, showing that the Mistral model show consistency in performance across the splits, albeit it is not the best model. Llama2-13B models perform the best for both the settings (definition and fulltext). The results also show that the performance difference between the definition and fulltext setting is marginal, as opposed to fine-tuned Language Models in Table \ref{tab:all_setting_comparison}.

\begin{table}[]
\resizebox{\columnwidth}{!}{%
\begin{tabular}{lrrrr}
\hline
\multicolumn{1}{c}{\multirow{2}{*}{\textbf{Model Name}}} & \multicolumn{2}{c}{\textbf{Zero-shot}}                                          & \multicolumn{2}{c}{\textbf{Few-shot}}                                           \\ \cline{2-5} 
\multicolumn{1}{c}{}                                     & \multicolumn{1}{c}{\textbf{Definition}} & \multicolumn{1}{c}{\textbf{Fulltext}} & \multicolumn{1}{c}{\textbf{Definition}} & \multicolumn{1}{c}{\textbf{Fulltext}} \\ \hline
Llama2-7B                                                & 0.48                                    & 0.50                                   & 0.51                                    & 0.52                                  \\
Llama2-13B                                               & 0.57                                    & 0.58                                   & 0.59                                    & 0.59                                  \\
Llama3-8B                                                & 0.33                                    & 0.40                                   & 0.35                                    & 0.40                                  \\
Mistral-7B                                               & 0.53                                    & 0.56                                   & 0.54                                    & 0.58                                  \\ \hline
\end{tabular}%
}
\caption{F1 score for Prompting experiment in Zero and Few shot setting for Definition and Fulltext splits}
\label{tab:f1_scores_instruct}
\end{table}

\noindent\textbf{Fine-tuning: } We also fine-tune the models tested with perplexity and instruction tuning with HoaxPedia with supervised fine-tuning \cite{touvron2023llama}. The results of fine-tuning as Macro-F1 scores for both definition and fulltext setting are shown in Table \ref{tab:f1_scores_finetuning}. The results show significant improvement across all the settings for all the models. Llama3 shows most consistency across the settings, and also is the best model across the scenarios, with a performance improvement of more than 25\% across the scenarios. However, the performance of the models in 1:100 hoax to legitimate scenario for all the models are around 50\% F1 in both settings. Further analysis reveals that the models only predict the majority label, which can further attributed to imbalance in the dataset \cite{powers2020evaluation}. A suitable weight adjusted training paradigm is required to handle the imbalance, but since it is not the scope of this work, we keep it for future work.

\begin{table}[]
\resizebox{\columnwidth}{!}{%
\begin{tabular}{lrrrrrr}
\hline
\multicolumn{1}{c}{\multirow{2}{*}{\textbf{Model}}} & \multicolumn{3}{c}{\textbf{Definition}}                                                                      & \multicolumn{3}{c}{\textbf{Fulltext}}                                                                        \\ \cline{2-7} 
\multicolumn{1}{c}{}                                & \multicolumn{1}{c}{\textbf{1H2R}} & \multicolumn{1}{c}{\textbf{1H10R}} & \multicolumn{1}{c}{\textbf{1H100R}} & \multicolumn{1}{c}{\textbf{1H2R}} & \multicolumn{1}{c}{\textbf{1H10R}} & \multicolumn{1}{c}{\textbf{1H100R}} \\ \hline
Llama2-7B                                           & 0.76                              & 0.47                               & 0.49                                & 0.66                              & 0.48                               & 0.47                                \\
Llama2-13B                                          & 0.80                              & 0.48                               & 0.50                                & 0.60                              & 0.63                               & 0.50                                \\
Llama3-8B                                           & 0.80                              & 0.48                               & 0.50                                & 0.83                              & 0.67                               & 0.50                                \\
Mistral-7B                                          & 0.71                              & 0.55                               & 0.49                                & 0.68                              & 0.53                               & 0.49                                \\ \hline
\end{tabular}%
}
\caption{Macro-F1 score for LLM Fine-tuning in degrees of data imbalance for Definition and Fulltext Settings}
\label{tab:f1_scores_finetuning}
\end{table}

\section{Conclusion and Future Work}

We have introduced \textsc{HoaxPedia}, a dataset containing hoax articles extracted from Wikipedia, from a number of sources, from official lists of hoaxes, existing datasets, and the Web Archive. We paired these hoax articles with similar legitimate articles, and after analyzing their main properties (concluding they are written with very similar style and content), we report the results of a number of binary classification experiments, where we explore the impact of (1) positive to negative ratio; and (2) going from the whole article to only the definition. This is different from previous work in that we have exclusively looked at the content of these hoax articles, rather than metadata such as traffic or longevity. For the future, we would like to further refine what are the criteria used by Wikipedia editors to detect hoax articles, and turn those insights into a ML model, and explore other types of non-obvious online vandalism.

\section{Limitations}
We present a new dataset named \textsc{Hoaxpedia} and associated baselines from a wide variety of language models / large language models in this work. Our study shows that these types of dataset can be helpful in the area of free text disinformation detection. However, there are some limitations to our work that we aim to address here. The sets proposed here are small, with only 311 positive examples (hoaxes), which can be attributed to the fact that we only collect the examples that are explicitly labeled as hoaxes, rather than articles under discussion for hoaxes. Additionally, in our experiments, we do not conduct further investigation for model behaviors such as performance improvement of Longformer models in the hardest setting (cf. Section \ref{sec:result}, implementing a weight-adjusted training paradigm for fine-tuning of LLMs (cf. Section \ref{sec:result_llm}) etc. We leave these analysis in future work, as the scope of this work is to introduce this dataset and establish the baseline results. Finally, we do not compare the results with existing work, mainly with \cite{kumar2016disinformation}, since the approaches mentioned in existing work are metadata dependent, and our approach is based on article text, we argue that the results may not be comparable.

\section{Ethics Statement}

This paper is in the area of online vandalism and disinformation detection, hence a sensitive topic. All data and code will be made publicly available to contribute to the advancement of the field. However, we acknowledge that deceitful content can be also used with malicious intents, and we will make it clear in any associated documentation that any dataset or model released as a result of this paper should be used for ensuring a more transparent and trustworthy Internet.

\bibliography{custom}

\appendix

\section{Dataset Details}
\label{app:dataset_details}
We release our dataset in 3 settings as mentioned in Section \ref{expt}. The settings with data splits and their corresponding sizes are mentioned in Table \ref{tab:dataset_details}.

\begin{table}[]
\resizebox{\columnwidth}{!}{%
\begin{tabular}{llllll}
\hline
                              &                             &                & \multicolumn{3}{l}{\textbf{Number of Instances}}   \\ \hline
\textbf{Dataset Setting}      & \textbf{Dataset Type}       & \textbf{Split} & \textbf{Non-hoax} & \textbf{Hoax} & \textbf{Total} \\ \hline
\multirow{2}{*}{1Hoax2legitimate}   & \multirow{2}{*}{Definition} & Train          & 426               & 206           & 632            \\ \cline{3-6} 
                              &                             & Test           & 179               & 93            & 272            \\ \hline
\multirow{2}{*}{1Hoax2legitimate}   & \multirow{2}{*}{Full Text}  & Train          & 456               & 232           & 688            \\ \cline{3-6} 
                              &                             & Test           & 200               & 96            & 296            \\ \hline
\multirow{2}{*}{1Hoax10legitimate}  & \multirow{2}{*}{Definition} & Train          & 2,225              & 203           & 2,428           \\ \cline{3-6} 
                              &                             & Test           & 940               & 104           & 1,044           \\ \hline
\multirow{2}{*}{1Hoax10legitimate}  & \multirow{2}{*}{Full Text}  & Train          & 2,306              & 218           & 2,524           \\ \cline{3-6} 
                              &                             & Test           & 973               & 110           & 1,083           \\ \hline
\multirow{2}{*}{1Hoax100legitimate} & \multirow{2}{*}{Definition} & Train          & 20,419             & 217           & 20,636          \\ \cline{3-6} 
                              &                             & Test           & 8,761              & 82            & 8,843           \\ \hline
\multirow{2}{*}{1Hoax100legitimate} & \multirow{2}{*}{Full Text}  & Train          & 22,274             & 222           & 22,496          \\ \cline{3-6} 
                              &                             & Test           & 9,534              & 106           & 9,640           \\ \hline
\end{tabular}%
}
\caption{Dataset details in different settings and splits}
\label{tab:dataset_details}
\end{table}

\section{Language Model Training Details}
\label{app:training_details}
We train our Language Models with the configuration given below. We use one NVIDIA RTX4090 to train the LMs, one NVIDIA V100 and one NVIDIA A100 GPU to train the LLMs.
\begin{itemize}
    \item Learning rate: 2e-06
    \item Batch size: 4 (for Fulltext experiments) and 8 (For Definition experiments)
    \item Epochs: 30
    \item Loss Function: Weighted Cross Entropy Loss
    \item Gradient Accumulation Steps: 4
    \item Warm-up steps: 100
\end{itemize}

\section{Prompt for LLM in-context learning}
\label{app:prompt_llm}
The instruction prompt used for LLMs in their in-context learning with examples for few shot experiment are given below.

\begin{tcolorbox}
\texttt{You are a helpful knowledge management expert and excel at identifying whether an input Wikipedia article is a hoax or not. Wikipedia defines a hoax as `a deliberately fabricated falsehood made to masquerade as truth'. You take an Wikipedia article as input and return with the label citing hoax(Label 1) or real(Label 0) based only on the text of the article. Given an article from Wikipedia, your task is to analyze the article text to identify if the article is hoax or real. The Hoax and real articles are defined as follows:}
\begin{itemize}
    \item \texttt{Hoax: An article that is deliberately fabricated falsehood made to masquerade as truth.}
    \item \texttt{Real: An article which contains information about an existing entity and are not fabricated.}
\end{itemize}

\texttt{Your output should be a JSON dictionary with label that you found. Here are the possible labels with what they mean:}
\begin{itemize}
    \item \texttt{0 : The article is real article.}
    \item \texttt{1 : The article is a hoax article.}
\end{itemize}

\texttt{Your input will be in the following format:}

\texttt{INPUT:}
\{
   \texttt{Text: <Article text>}
\}

\texttt{OUTPUT:}
    \{
        \texttt{Label: <One of the label from the possible labels: 0 and 1, where 0 is real article and 1 is hoax article.>}
    \}
    
\texttt{Please respond with only the JSON dictionary containing label. You are instructed strictly to return output only in the format given above, nothing else. No yapping.}
\end{tcolorbox}

Here are the examples used in few-shot experiments.

\begin{tcolorbox}
\textbf{Example 1:}

\texttt{INPUT:}

\{
    \texttt{Text:  Albion Dauti (born May 31, 1995 in Caracas, Venezuela) is a Venezuelan telenovela actor and presenter. Albion Dauti was born in Caracas, Venezuela. He studied acting at the Faculty of Arts in Caracas. In 2010 he started working as a presenter in Venevision.}
\}

\texttt{OUTPUT:}

\{
    \texttt{Label: 1}
\}

\textbf{Example 2:}

\texttt{INPUT:}

\{
    \texttt{Text:}\texttt{ Michelle Madhok (born May 26, 1971) is the Founder and CEO of White Cat Media Inc. - DBA SheFinds Media, parent company of online shopping publication SheFinds. com and MomFinds. com. She writes a weekly style column for New York's Metro newspaper and appears regularly on Fox News Channel, The Today Show, and The Tyra Banks Show.}
\}

\texttt{OUTPUT:}

\{
    \texttt{Label: 0}
\}
\end{tcolorbox}

\end{document}